\documentclass[default]{sn-jnl}

\jyear{2022}%

\theoremstyle{thmstyleone}%
\theoremstyle{thmstyletwo}%

\theoremstyle{thmstylethree}%
\begin{document}

\title[]{Spatio-Temporal-based Context Fusion for Video Anomaly Detection}


\author{\fnm{Chao} \sur{Hu}}
\author{\fnm{Weibin} \sur{Qiu}}
\author{\fnm{Weijie} \sur{Wu}}
\author{\fnm{Liqiang} \sur{Zhu}}

\affil{\orgdiv{AI Lab}, \orgname{Unicom (Shanghai) Industry Internet Co., Ltd.}, \orgaddress{\city{Shanghai} \postcode{200000},  \country{China}}}


\abstract{Video anomaly detection aims to discover abnormal events in videos, and the principal objects are target objects such as people and vehicles. Each target in the video data has rich spatio-temporal context information. Most existing methods only focus on the temporal context, ignoring the role of the spatial context in anomaly detection. The spatial context information represents the relationship between the detection target and surrounding targets. Anomaly detection makes a lot of sense. To this end, a video anomaly detection algorithm based on target spatio-temporal context fusion is proposed. Firstly, the target in the video frame is extracted through the target detection network to reduce background interference. Then the optical flow map of two adjacent frames is calculated. Motion features are used multiple targets in the video frame to construct spatial context simultaneously, re-encoding the target appearance and motion features, and finally reconstructing the above features through the spatio-temporal dual-stream network, and using the reconstruction error to represent the abnormal score. The algorithm achieves frame-level AUCs of 98.5\% and 86.3\% on the UCSDped2 and Avenue datasets, respectively. On the UCSDped2 dataset, the spatio-temporal dual-stream network improves frames by 5.1\% and 0.3\%, respectively, compared to the temporal and spatial stream networks. After using spatial context encoding, the frame-level AUC is enhanced by 1\%, which verifies the method's effectiveness.}

\keywords{Anomaly Detection, context fusion, optical flow, spatial context, temporal context, dual-stream network}



\maketitle

\section{INTRODUCTION}\label{sec1}

With the economic and social development, video surveillance systems are more and more widely used in public places and play an important role in maintaining social security and stability. Video anomaly detection refers to the detection of unexpected or common behavior in video through algorithms, aiming to find and locate abnormal behaviors that may threaten public safety, with a very broad application scenario. There are two main difficulties of video anomaly detection. On the one hand, abnormal events have strong scene dependence, different scenes have different definitions of abnormal behavior, and normal events in some scenes will become abnormal in other scenes, such as normal when people walk on the sidewalk, but crossing the road at will is abnormal. On the other hand, abnormal events have the characteristics of scarcity, diversity, and inexhaustibility. Therefore, video anomaly detection mostly uses semi-supervised or unsupervised methods, using a training set containing only normal samples to train models and detect test sets.

The main body of the abnormal event is often the pedestrian, vehicle and other targets in the video, and most of the existing anomaly detection methods divide the video frame into several intervals for processing or directly input the video frame into the model, which cannot pay good attention to the abnormal object and is susceptible to background interference. In terms of detection methods, most of the existing algorithms only consider time context extraction and ignore the spatial context information, which is of great significance to anomaly detection for targets in continuous video data. The temporal context information represents the motion state of the target, and the spatial context information represents the relationship between the detected target and the surrounding target. For example, in a test set video frame, if both a pedestrian and a car appear at the same time, but this does not occur in the training set, an abnormal target can be considered to have occurred.

Aiming at these problems, we propose an anomaly detection algorithm based on the fusion of target spatio-temporal contexts. Target extraction is employed to avoid background interference, while building spatial context with multiple targets in the same video frame and re-encoding target appearance and motion characteristics. The appearance and motion characteristics of the target are reconstructed separately through the spatio-temporal dual-flow network, and the two subnetworks adopt the same structure, including the autoencoder and the context coding memory module, which takes the reconstruction error as the anomaly score, and makes full use of the temporal and spatial context information of the target to more accurately detect the abnormal target.

\section{RELATED WORK}\label{sec2}

Video anomaly detection algorithms can be divided into two categories: traditional machine learning methods and deep learning methods. Traditional machine learning methods are generally divided into two steps: feature extraction and anomaly discrimination. The extracted manual features include: optical flow histogram \cite{ref3}, texture \cite{ref4}, 3D gradient \cite{ref5}, etc. Commonly used discrimination methods include: clustering discrimination \cite{ref6}, reconstruction discrimination \cite{ref7}, and generation probability discrimination \cite{ref8}. These handcrafted feature-based methods are often difficult to describe complex objects and behaviors in video anomaly recognition, and the results are not ideal.

In the past ten years, deep learning-based methods have been widely used in surveillance video anomaly detection and have made great progress. For the extraction of spatiotemporal information, there are mainly two-stream network, 3D convolution and LSTM methods. Karen S et al. \cite{ref9} proposed a spatio-temporal dual-stream network to train two CNN models for video images and optical flow respectively, and finally merge the discriminative results. Fan et al. \cite{ref10} used a dual-stream network for video anomaly recognition, using RGB frames and dynamic streams to represent appearance anomalies and motion anomalies, respectively, and used Gaussian mixture models for anomaly discrimination. 3D convolution is also a common method, and there are various 3D convolution models such as C3D, I3D, and S3D \cite{ref11}. Nogas J et al. \cite{ref12} used a 3D convolutional autoencoder to calculate the reconstruction error of video spatio-temporally fast to discriminate anomalies. The network structure is simple and efficient, but the results are often not ideal. People also use LSTM or RNN cyclic neural network to model continuous video frames. Chong Y S et al. \cite{ref13} used LSTM to extract video features, and then used AE to reconstruct the features and discriminate abnormalities. The network structure is relatively complex and the training cost is high. This paper adopt dual-stream network structure.

Commonly used video anomaly detection networks in deep learning include VAE, memAE, GAN, etc. Wang et al. \cite{ref14} proposed a concatenated VAE structure, filtering apparently normal samples through the first VAE network, and then using the second VAE network for anomaly detection. Sabokrou M et al. \cite{ref15} combined CAE and GAN, and used the discriminator of GAN to improve the reconstruction ability of CAE. However, due to the strong generalization ability of the autoencoder, abnormal samples may also obtain lower reconstruction errors. In order to increase the reconstruction error of abnormal samples, memAE is used for video anomaly detection. GongD et al. \cite{ref16} in the AE method added the Memory module, which reduces the generalization ability of the model and makes abnormal samples have larger reconstruction errors. Park H et al. \cite{ref17} added feature separation loss which based on memAE, that learned multiple modes of normal video frames, and used U-net instead of encoder, and realized reconstruction and prediction at the same time. Most of these methods divide the video frame into several intervals for processing or directly input the video frame into the model, which cannot pay attention to the abnormal objects and are easily disturbed by the background. For this reason, Ionescu RT et al. \cite{ref18} regard video anomaly detection as a classification problem. , the target in the video frame is separated by target detection, and then K-mean and SVM are used to classify the target. When the target does not belong to any category, it is detected as an abnormal object. However, this method regards the extracted targets as separate individuals, ignoring the relationship between the targets and the surrounding targets in space, so some context-related anomalies cannot be well detected.

\section{THE PROPOSED ANOMALY DETECTION FRAMEWORK}\label{sec3}

\begin{figure*}
    \centering
    \includegraphics[width=1.0\textwidth]{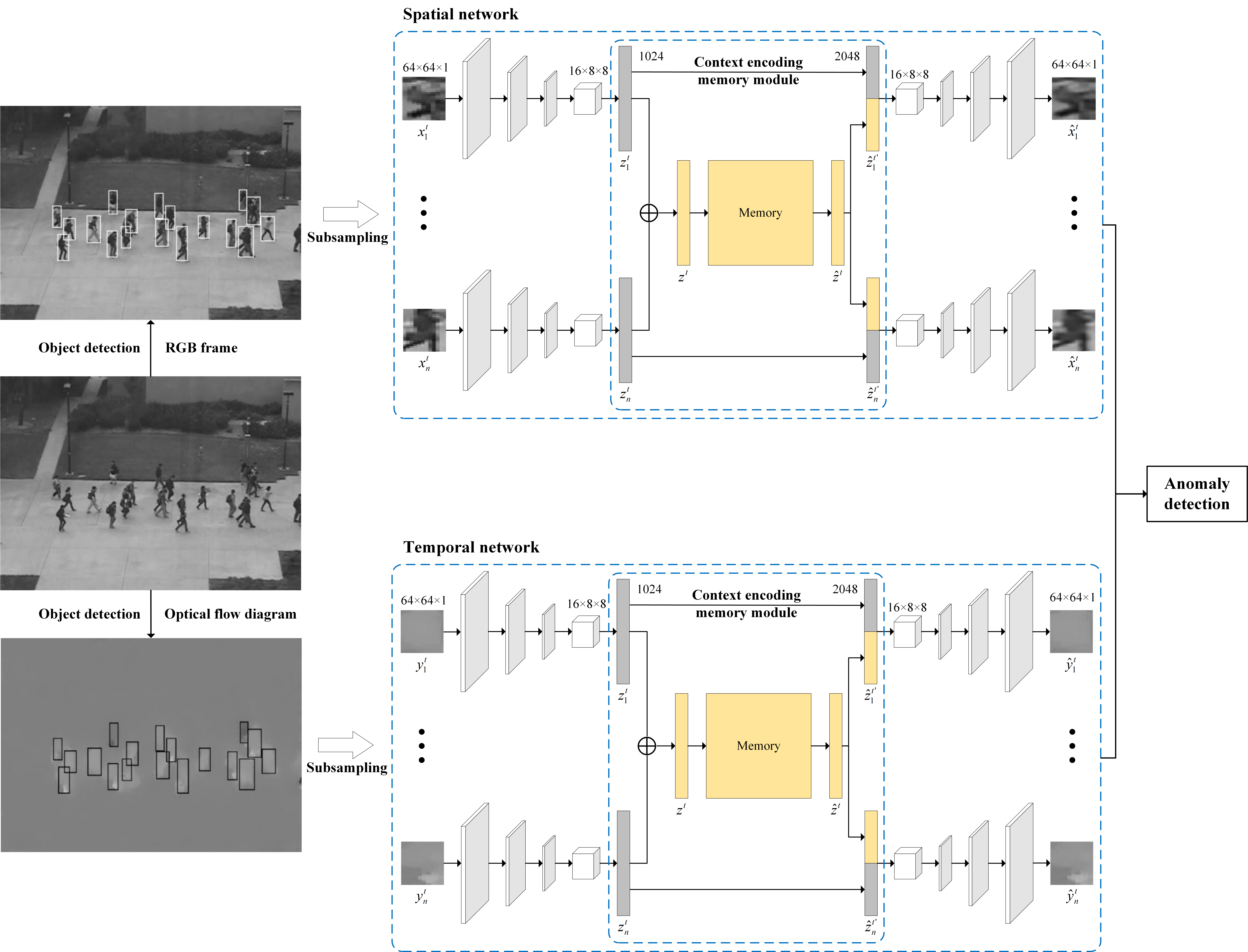}
    \caption{\centering{Network Framework.}}
    \label{fig1}
    
\end{figure*}

The network framework of this paper is composed of target detection module, temporal network, spatial network and anomaly detection module. The two sub-networks consist of an encoder, a context encoding memory module and a decoder, respectively. For the video data set, firstly divide the frame and perform target detection on the video frame to obtain the rectangular bounding box of all targets, then calculate the optical flow, and extract the moon marker at the corresponding position of the video frame and the optical flow map, and input the spatial network and temporal network respectively. Spatial network extracts appearance features, and temporal network extracts motion features. After the target is encoded, context extracted, and reconstructed, the reconstruction error is calculated. As anomaly score, the anomaly detection module fuses the outputs of the two sub-networks and gives the final anomaly evaluation.

\subsection{Proposed Method}\label{subsec3.1}
This paper uses the FPN \cite{ref19} network for object detection, which combines both the underlying detail information and the high-level semantic information, which can detect smaller targets well, while taking into account accuracy and speed. For the video dataset $\mathit{X}$, first get the continuous video frame $\mathit{X}^1, \mathit{X}^2, \ldots, \mathit{X}^\mathit{N}$ by frame, enter the FPN network in turn, get the rectangular bounding box of all the targets in the video, and then calculate the optical flow graph of the adjacent two frames:

\begin{equation}
\mathit{Y}^\mathit{t}=f(\mathit{X}^{t-1}, \mathit{X}^t),
\label{equation:1}
\end{equation}

where the optical flow diagram $\mathit{Y}^\mathit{t}$ is calculated from the video frames $\mathit{X}^\mathit{t-1}$ and $\mathit{X}^\mathit{t}$. Then crop all the targets in the video frame at the corresponding positions on the RGB frame $\mathit{X}^\mathit{t}$ and the optical flow diagram $\mathit{Y}^\mathit{t}$ respectively, for each frame $\mathit{X}^\mathit{t}$, after the target detection is obtained, the target RGB frame $\mathit{x}^\mathit{1}, \mathit{x}^\mathit{2}, \ldots, \mathit{x}^\mathit{n}$ and the target optical flow diagram $\mathit{y}^\mathit{1}, \mathit{y}^\mathit{2}, \ldots, \mathit{y}^\mathit{n}$, zoomed to a matrix of size $64\times64$, in frames, the above objects are input into the spatial and time information network respectively.

\subsection{Spatial Network}\label{subsec3.2}

The function of the spatial network is to extract the appearance characteristics of the target object, and then decode and reconstruct it after the spatial context encoding, so as to determine whether the video frame has an abnormal object in the spatial domain.

\subsubsection{Spatial AutoEncoder}\label{subsubsec3.2.1}

The encoder encodes the input RGB frame as a feature vector, the input as the RGB frame $\mathit{x}^\mathit{t}_\mathit{i}$ of a single target, the output as the appearance feature $\mathit{z}^\mathit{t}_\mathit{i}$ of the target, and the encoder contains four convolutional layers and three maximum pooling layers, using the $\mathit{RELU}$ activation function. The decoder reconstructs the recoded feature to calculate the exception score, the input of the decoder is the recoded feature $\hat{z}^\mathit{t}_\mathit{i}$, the output is the reconstructed target $\hat{x}^\mathit{t}_\mathit{i}$, containing four convolutional layers and three upsampled layers, using the nearest neighbor interpolation method for upsampling.

\begin{equation}
\mathit{z}^\mathit{t}_\mathit{i} = \mathit{E}(\mathit{x}^\mathit{t}_\mathit{i},\theta_\mathit{e}),
\label{equation:2}
\end{equation}

\begin{equation}
\hat{x}^\mathit{t}_\mathit{i} = \mathit{D}(\mathit{x}^\mathit{t}_\mathit{i},\theta_\mathit{d}),
\label{equation:3}
\end{equation}

where $\theta_\mathit{e}$, $\theta_\mathit{d}$ represents the parameters of the encoder and decoder, $\mathit{E}$ represents the encoder, and $\mathit{D}$ represents the decoder.

\subsubsection{Context-Encoded Memory}\label{subsubsec3.2.2}

Through target extraction, the model only pays attention to the target that may be abnormal, thereby avoiding the interference of the background image, but at the same time it also removes the spatial context information of the target, that is, the relationship between multiple targets in the same frame, such as the co-occurrence relationship between vehicles and pedestrians on the playground.

To this end, we embed a contextual encoding memory module in the self-encoder, hoping to use the spatial context relationship to encode the target features and assist in target reconstruction, thereby increasing the reconstruction error of abnormal targets. The context encoding memory module input is the encoded target appearance feature $\mathit{z}^\mathit{t}_\mathit{i}$, and the features $\mathit{z}^\mathit{t}_\mathit{1}, \mathit{z}^\mathit{t}_\mathit{2}, \ldots, \mathit{z}^\mathit{t}_\mathit{n}$ of all targets in each frame are input at a time, which is used to calculate the spatial context of the frame and re-encode the features of these $\mathit{n}$ targets. First of all, the input feature map is expanded to obtain the feature vector of $\mathit{C}$ dimension, we believe that there is an association between different targets in the same frame, each frame of video can get $\mathit{n}$ targets, and after superimposing them to obtain a feature vector of size $\mathit{C}$, as the spatial context information of this video frame:

\begin{equation}
\mathit{z}^\mathit{t} = \frac{1}{n} \sum^{n}_{i-1}{\mathit{z}^\mathit{t}_\mathit{i}},
\label{equation:4}
\end{equation}

where $\mathit{z}^\mathit{t}$ represents the spatial context information of the video frame, $\mathit{z}^\mathit{t}_\mathit{i}$ represents the feature of the target $\mathit{x}^\mathit{t}_\mathit{i}$, and $\mathit{n}$ is the number of targets in a frame, if there are fewer than $\mathit{n}$ targets in a frame, add the targets to $\mathit{n}$ by random repeating.

\begin{figure*}
    \centering
    \includegraphics[width=0.9\textwidth]{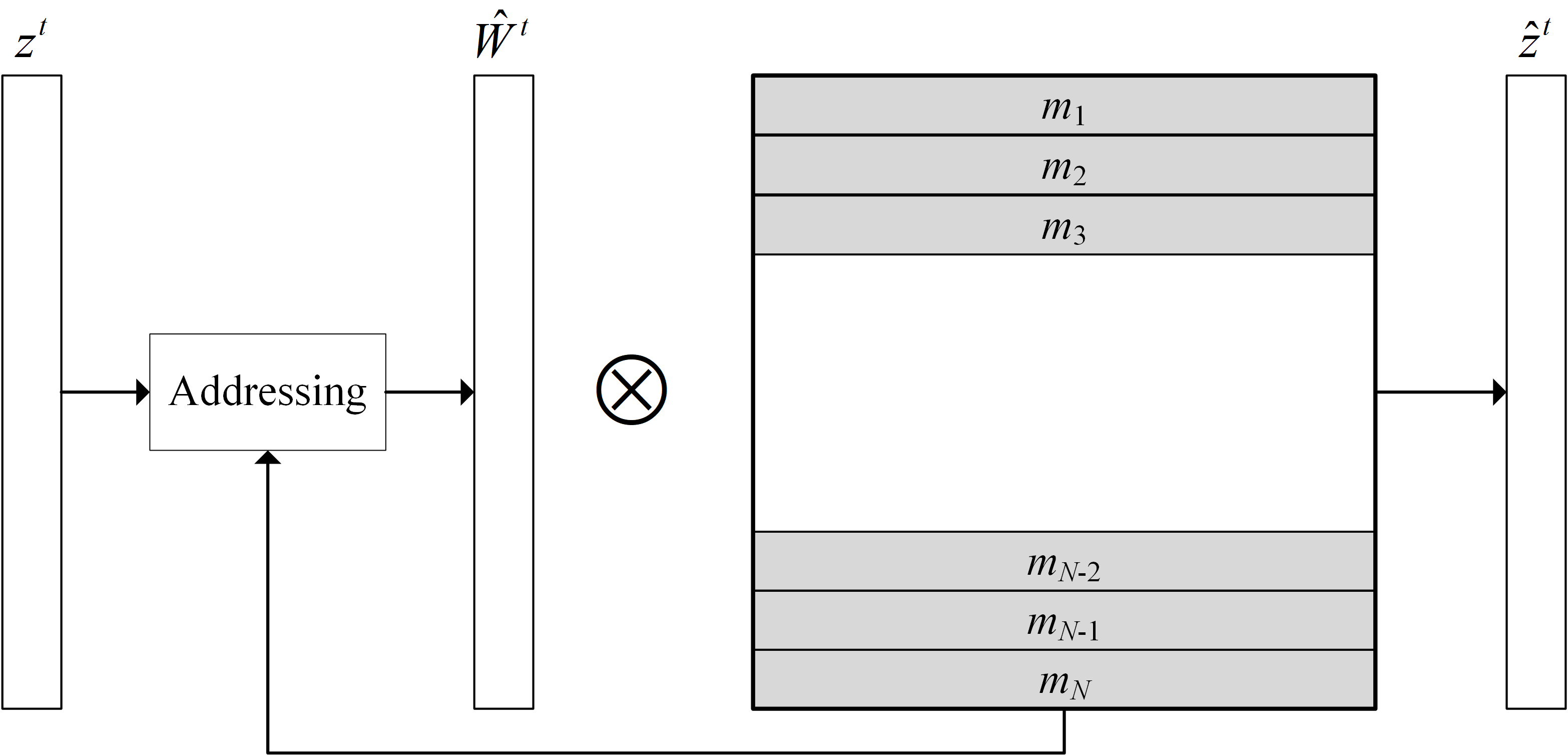}
    \caption{\centering{Memory network}.}
    \label{fig2}
\end{figure*}

Video anomaly detection uses unsupervised training, the training set data is all normal samples, and the spatial context information of each frame represents a normal mode, so the normal mode from the training set is the key to our anomaly detection of the test set. Due to the strong generalization ability of the autoencoder, so that part of the exception context information can also be better used for reconstruction, we added a memory module \cite{ref16} after the context encoding, aiming to learn a limited number of prototype features that can best characterize all normal context information, and save them in memory items, the structure is as shown in Figure \ref{fig2}, the memory block is a matrix of size $N\times{C}$, $M\in{R^{N\times{C}}} $, storing $\mathit{N}$ memory items, each memory item is $\mathit{m}_j$, the dimension is $\mathit{C}$, update and read the memory item during training, Only memory entries are read during testing.

The input of the memory module is the feature vector $\mathit{z}^\mathit{t}$ representing the spatial context information, and the output $\hat{x}^\mathit{t}$ is multiplied by the weight vector and the memory term. For the read operation of the feature $\mathit{z}^\mathit{t}$, first address, calculate the cosine similarity of $\mathit{z}^\mathit{t}$ and each memory term $\mathit{m}_j$ respectively, and then use the $\mathit{Softmax}$ function to calculate the weight matrix $\mathit{W}^\mathit{t}$:

\begin{equation}
\mathit{w}^\mathit{t}_\mathit{j} = \frac{\exp{(d(z^t,m_j))}}{\sum^{N}_{k=1}{\exp{(d(z^t,m_k))}}}
\label{equation:5}
\end{equation}

where $\mathit{d}(\cdot,\cdot)$ represents cosine similarity:

\begin{equation}
d(z^t,m_j) = \frac{{z^t}{m^T_j}}{\parallel z^t\parallel\parallel m_j\parallel  }
\label{equation:6}
\end{equation}

Exception context information may also be obtained from some normal feature combinations, in order to limit the reconstruction of exception context information, we do a hard shrinkage of the weight matrix $\mathit{W}^\mathit{t}$ to ensure its sparsity:

\begin{equation}
\hat{w}^\mathit{t}_\mathit{j} = h(w^t_j;\lambda) = \begin{cases}{w^t_i,}&{if(w^t_i>\lambda)}\\ {0,}&{otherwise}\end{cases} 
\label{equation:7}
\end{equation}

where $\lambda$ is the sparse threshold, and the weight becomes $0$ when it falls below the threshold, and hard compression prompts the model to use fewer memory terms to reconstruct the input features. $\hat{w}^\mathit{t}_\mathit{j}$ as the weight of the memory item $\mathit{m}_\mathit{j}$, for the feature $\mathit{z}^\mathit{t}$, the higher its similarity to the memory item $\mathit{m}_\mathit{j}$, the greater the weight, the read operation can be expressed as:

\begin{equation}
\hat{z}^\mathit{t} = \hat{W}^\mathit{t}M = \sum^{N}_{j=1}{\hat{w}^\mathit{t}{m_j}}
\label{equation:8}
\end{equation}

where $\hat{W}$ is the weight matrix corresponding to feature $\mathit{z}^\mathit{t}$ and $\mathit{M}$ is the memory term.

For the context information $\mathit{Z}^\mathit{t}$ calculated in Equation \ref{equation:3}, $\hat{Z}^\mathit{t}$ is obtained by reading memory, and then spliced with the feature $\mathit{Z}^\mathit{t}_\mathit{i}$ of each target to obtain the feature vector $\hat{Z}^{\mathit{t}^\prime}_\mathit{i}$ of $2 \times\mathit{C}$ dimension:

\begin{equation}
\hat{Z}^{\mathit{t}^\prime}_\mathit{i} = \mathit{Z}^\mathit{t}_\mathit{i} \bigoplus \hat{Z}^\mathit{t}
\label{equation:9}
\end{equation}

In this way, $\hat{Z}^{\mathit{t}^\prime}_\mathit{i}$ contains both the characteristics of the target itself and the spatial context information as input to the decoder for the next refactoring.

\subsubsection{Loss Function}\label{subsubsec3.2.3}

In this paper, the temporal network and the spatial network are trained separately, and the loss function of the spatial network consists of two parts: the loss of appearance characteristics and the loss of entropy. First, minimize the reconstruction error of each target, as an appearance feature loss, using the mean squared difference loss function:

\begin{equation}
L^{a}_{recon}(x^t_i)=\parallel x^t_i-\hat{x}^t_i\parallel 
\label{equation:10}
\end{equation}

In the memory module, in order to make the encoded spatial context feature $\mathit{Z}^\mathit{t}$ as similar as possible to the most similar one in memory, we added an entropy loss function:

\begin{equation}
L^{a}_{ent}(x^t)=\sum^{N}_{j-1}{-\hat{w}^t_j\log{(\hat{w}^t_j)}}
\label{equation:11}
\end{equation}

where $\hat{w}^t_j$ is the memory addressing weight in Equation \ref{equation:7}, and the two loss functions are balanced to obtain the total loss function:

\begin{equation}
L^{a}=\lambda_{recon}{L^a_{recon}} + \lambda_{ent}{L^a_{ent}}
\label{equation:12}
\end{equation}

where $\lambda_{recon}$ and $\lambda_{ent}$ represent the weights of reconstruction loss and entropy loss, respectively.

\subsection{Temporal Network}\label{subsec3.3}

This paper uses optical flow to represent the motion information of the target, and the optical flow has the invariance of apparent characteristics, which not only represents the motion characteristics of the target, but also eliminates the interference of color and light intensity on anomaly detection, which can well make up for the lack of RGB frames \cite{ref20}. The inputs of the temporal network are the target optical flow graphs $\mathit{y}^t_1, \mathit{y}^t_2, \ldots, \mathit{y}^t_n$ for each frame, and the output is their recomposition pictures $\mathit{y}^t_1, \mathit{y}^t_2, \ldots, \mathit{y}^t_n$. The structure of the temporal network is the same as that of the spatial network, and the loss function consists of optical flow characteristic loss and entropy loss, similar to Equation \ref{equation:12}:

\begin{equation}
L^{m}=\lambda_{recon}{L^m_{recon}} + \lambda_{ent}{L^m_{ent}}
\label{equation:13}
\end{equation}

\subsection{Anomaly Detection}\label{subsec3.4}

The data set $\mathit{X}$ is first framed to obtain continuous video frames $\mathit{X}^1, \mathit{X}^2, \ldots, \mathit{X}^N$, target extraction and optical flow calculation for each frame $\mathit{X}^t$, and multiple target RGB frames $\mathit{x}^1, \mathit{x}^2, \ldots, \mathit{x}^n$ and optical flow graphs $\mathit{y}^1, \mathit{y}^2, \ldots, \mathit{y}^n$, respectively, input spatio-temporal information network, and obtain the reconstruction error of RGB frame and optical flow and regularization according to Equation \ref{equation:10}:

\begin{equation}
\hat{L}^a_{recon}(x^t_i)=\frac{L^a_{recon}(x^t_i)-\min{(L^a_{recon}(x^t_i))}}{\max{(L^a_{recon}(x^t_i))}}
\label{equation:14}
\end{equation}

\begin{equation}
\hat{L}^m_{recon}(y^t_i)=\frac{L^m_{recon}(y^t_i)-\min{(L^m_{recon}(y^t_i))}}{\max{(L^m_{recon}(y^t_i))}}
\label{equation:15}
\end{equation}

$\hat{L}^a_{recon}(x^t_i)$ and $\hat{L}^m_{recon}(x^t_i)$ represent the RGB frame and optical flow reconstruction error after regularization, and $\max{(L^s_{recon}(x^t_i))}$ and $\max{(L^t_{recon}(x^t_i))}$ represent the maximum value of RGB frame reconstruction error and optical flow reconstruction error in the entire dataset of targets.

Then, the joint reconstruction error $\hat{L}_{recon}(x^t,y^t)$ for each target is calculated, the RGB frame reconstruction error represents the appearance anomaly of the target, and the optical flow reconstruction error represents the motion anomaly of the target, which we believe is abnormal when one of the appearance and motion states is abnormal. Therefore, the maximum value of the appearance anomaly and the movement anomaly of each target is taken as the anomaly score for that target:

\begin{equation}
\hat{L}_{recon}(x^t_i, y^t_i) = \max{(\hat{L}^a_{recon}(x^t_i), \hat{L}^m_{recon}(y^t_i))}
\label{equation:16}
\end{equation}

Joint reconstruction error $\hat{L}_{recon}(x^t_i, y^t_i)$ takes the maximum value of RGB frame reconstruction error and optical flow reconstruction error, and after obtaining the abnormal score of each target, the abnormal score of all targets in the video frame is maximized, as the abnormal score $\mathit{S(X^t)}$ of sample $\mathit{X^t}$, that is, if an abnormal target appears in a frame, it is judged to be an abnormal frame:

\begin{equation}
S(X^t) = \max{(\hat{L}^a_{recon}(x^t_i,y^t_i))},\forall{i\in\{1,2,\ldots,n\}}
\label{equation:17}
\end{equation}

Finally, the adjacent 10-frame video anomaly scores are averaged for fractional smoothing.

\section{EXPERIMENT}\label{sec4}

\subsection{DataSet}\label{subsec4.1}

This paper uses the USCD and Avenue datasets, the UCSD \cite{ref21} pedestrian dataset contains two subsets of ped1 and ped2, this paper uses the ped2 subset, the training set contains 16 videos for a total of 2550 frames, and the test set contains 11 videos for a total of 2010 frames with $240\times360$ pixels per frame, and abnormal behavior includes bicycles, cars, and skateboards. The Avenue \cite{ref22} dataset training set contains 16 videos with a total of 15328 frames, and the test set contains 21 videos with a total of 15324 frames with $360\times640$ pixels per frame, and abnormal events include running, wrong walking direction, cycling, etc.

\subsection{Parameter Settings And Evaluation Metrics}\label{subsec4.2}

This paper uses the ResNet50fpn object detection network pre-trained on coco data to extract targets, for the training set and test set, the detection thresholds are set to 0.5 and 0.4 respectively, to reduce the generalization ability of the network, expand the reconstruction loss, the number of targets n is set to 18 and 24 respectively, TVL1 extracts optical stream, all video frames are converted to grayscale pictures, and the captured targets are scaled to $64\times64$ pixels. The autoencoder adopts Adam optimizer optimization, and sets the learning rate of 0.001 and 0.0001 for the spatial information network and the time information network respectively, with a batch size of 64, hyperparameters $\lambda_{recon}=1.0$, $\lambda_{ent}=0.0002$. In this paper, the area under the area under the curve (AUC) of the subject's operating characteristic curve (ROC) is used as the evaluation index, and the higher the AUC, the better the anomaly detection effect.

\subsection{Expriment Results}\label{subsec4.3}

\subsubsection{Different Thresholds For FPN Networks}\label{subsubsec4.3.1}

The detection threshold of the FPN network determines how many targets can be extracted, too low the threshold can lead to missed detection, and too high the threshold may misjudge the background as a target. On the UCSD-ped2 dataset, we experimented with the effects of different thresholds of FPN networks on anomaly detection, and the results are shown in Table \ref{tab1}.

\begin{table}[!ht]
\begin{center}
\setlength{\tabcolsep}{6.5mm}
\caption{\centering{Inference of FPN confidence level AUC(\%)}}\label{tab1}
\begin{tabular}{cc}
\toprule
Threshold(Training/Test) & UCSD-ped2\\
\midrule
0.4/0.4    &    97.9  \\
0.4/0.5    &    95.5  \\
0.5/0.5    &    95.4  \\
0.5/0.4    &    98.5 \\
\botrule
\end{tabular}
\end{center}
\end{table}

This gives the best results from setting the detection thresholds for the training set and the test set to 0.5 and 0.4, respectively. For training sets, higher detection thresholds can obtain fewer targets, reducing the generalization ability of the network, increasing the reconstruction error, and for test sets, when the detection threshold is low, some abnormal targets will be missed, affecting the results.

\subsubsection{Comparison With Other Models}\label{subsubsec4.3.2}

This paper compared the method in this paper with some existing video anomaly detection algorithms on the frame-level AUC, and the results are shown in Table \ref{tab2}.

\begin{table}[!ht]
\center
\setlength{\tabcolsep}{6mm}
\caption{\centering{Comparison of abnormal detection methods AUC(\%)}}\label{tab2}
\begin{tabular}{ccc}
\toprule
\textbf{Method} & \textbf{UCSD-ped2} & \textbf{Avenue}\\
\midrule
Conv-AE \cite{ref23}    &            90.0            &   70.2      \\
ConvLSTM-AE \cite{ref24}    &        88.1        &    77.0      \\
Mem-AE \cite{ref16}    &        95.4        &    84.9      \\
GMFC-VAE \cite{ref10}    &        92.2    &    83.4      \\
FPN-AE-SVM \cite{ref18}    &        97.0        &    88.5      \\
P w/ MemAE \cite{ref17}    &        97.0        &    88.5      \\
AMC \cite{ref25}    &        96.2        &    86.9      \\
\midrule
\textbf{Ours}    &        \textbf{98.5}        &        \textbf{86.3}      \\
\botrule
\end{tabular}
\end{table}

It can be seen that our method has achieved the best results on the UCSD-ped2 dataset, compared with the same object detection algorithm FPN-AE-SVM \cite{ref18}, our results are improved by 1.5\%, compared with the algorithms Mem-AE \cite{ref16} and P w/ MemAE \cite{ref17} using memory modules, our results are also greatly improved, because there are more targets per frame in this data set, and the abnormal events are a small number of bicycles and cars, skateboards, and the normal mode can be well learned through spatial context information fusion to check out the abnormal target. Our method also achieves good results on the Avenue dataset.

\begin{figure*}
    \centering
    \includegraphics[width=0.8\textwidth]{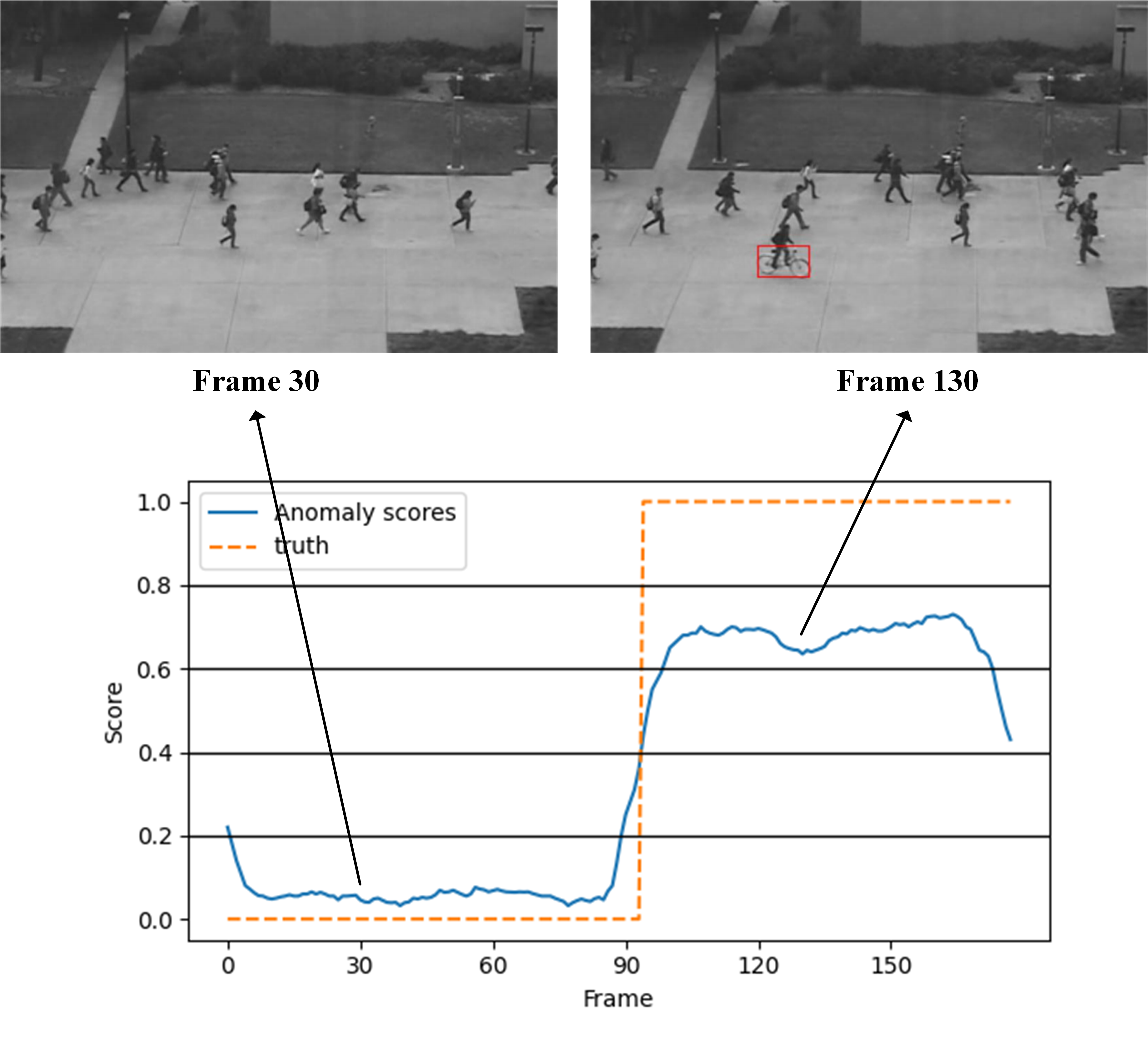}
    \caption{\centering{scores of UCSD-ped2 dataset}.}
    \label{fig3}
\end{figure*}

Figure \ref{fig3} shows the exception score and real label on UCSD-ped2 Test Set Video II.

\subsection{Ablation Study}\label{subsec4.4}

In order to verify the role of the dual-stream network and context-encoded memory modules, we designed a comparative experiment on the UCSD-ped2 dataset.

\begin{table}[!ht]
\center
\setlength{\tabcolsep}{6mm}
\caption{\centering{Comparison of spatio-temproal network and context fusion AUC(\%)}}\label{tab3}
\begin{tabular}{ccc}
\toprule
\textbf{Method} & \textbf{Context-encoded memory} & \textbf{UCSD-ped2} \\
\midrule
Spatial Network    &    &  92.4\\
Temporal Network    &       & 97.2 \\
Dual Network    &    & 97.5   \\
Spatial Network &  \checkmark &        92.7   \\
Temporal Network &  \checkmark  &        98.2     \\
Dual Network &  \checkmark &        98.5   \\
\botrule
\end{tabular}
\end{table}

Table \ref{tab3} show the temproal network based on the optical flow characteristics is better than the spatial network, and the Spatio-temporal dual network makes full use of the temporal and spatial network in the video, which is 5.1\% and 0.3\% higher than the spatial and temporal network, respectively. After adding the context-encoded memory module, the frame-level AUC of all three networks has been improved, and the AUC of the dual network has increased by 1\%, which proves the role of context-encoded memory module in video anomaly detection.

\section{Conclusions}\label{sec5}

In this paper, a video anomaly detection method based on target spatio-temporal context fusion is proposed, which reduces background interference through target extraction. Simultaneously building spatial context with multiple targets in video frames and re-encoding target appearance and motion features makes full use of the target's spatial and temporal context information, and experiments on USCD-ped2 and Avenue datasets verify the effectiveness of the method. Since the object detection algorithm will produce a certain amount of missed detection, resulting in abnormal target misjudgment, the selection of the object detection algorithm will be optimized in the future, and the construction method of improving the spatial context for the connection between the targets will also be considered.

\end{document}